\documentclass{article}
\usepackage[utf8]{inputenc} %
\usepackage[T1]{fontenc}    %
\usepackage{url}            %
\usepackage{spconf,amsmath,graphicx}
\usepackage{booktabs}       %
\usepackage{amsfonts}       %
\usepackage{nicefrac}       %
\usepackage{microtype}      %
\usepackage{multirow}  
\usepackage{titlesec} %
\usepackage{hyperref} %
\usepackage{adjustbox}  
\usepackage{rotating}   
\usepackage{amsmath}
\usepackage{enumitem}
\usepackage{xcolor}
\usepackage{arydshln}
\usepackage{cleveref}  %

\title{TrendGNN: Towards Understanding of Epidemics, Beliefs, and Behaviors}
\name{Mulin Tian, Ajitesh Srivastava}
\address{University of Southern California}
\begin{document}
\maketitle
\begin{abstract}
Epidemic outcomes have a complex interplay with human behavior and beliefs. Most of the forecasting literature has focused on the task of predicting epidemic signals using simple mechanistic models or black-box models, such as deep transformers, that ingest all available signals without offering interpretability. However, to better understand the mechanisms and predict the impact of interventions, we need the ability to forecast signals associated with beliefs and behaviors in an interpretable manner. In this work, we propose a graph-based forecasting framework that first constructs a graph of interrelated signals based on trend similarity, and then applies graph neural networks (GNNs) for prediction. This approach enables interpretable analysis by revealing which signals are more predictable and which relationships contribute most to forecasting accuracy. We believe our method provides early steps towards a framework for interpretable modeling in domains with multiple potentially interdependent signals, with implications for building future simulation models that integrate behavior, beliefs, and observations.
\end{abstract}


\section{Introduction}

Understanding the interplay between behaviors, beliefs, and epidemic outcomes remains a fundamental challenge, requiring comprehensive models to guide effective public health interventions. Capturing these dynamics is particularly difficult, as the signals are high-dimensional, noisy, and interdependent across regions and populations. However, many existing models rely on overly simplistic assumptions about these mechanisms ~\cite{suchoski2022gpu,linas2022projecting,davis2021cryptic,lemaitre2021scenario,chen2021epidemiological}. Further, existing forecasting methods~\cite{bracher2021evaluating,ray_ensemble_2020} are focused on epidemic signals and do not provide any insight into how these might influence (or predict) behaviors and beliefs, and vice versa.

In this work, we present a graph-based framework that models these relationships explicitly. We construct graphs to represent dependencies among features and states, and then apply GraphSAGE to leverage these structured relationships for multi-week forecasts. Our study shows that both the use of graph neural networks and the choice of graph construction strategy are critical for improving mid- to long-term forecasting. Beyond accuracy, the graph-based perspective provides interpretability by highlighting which signals and dependencies matter most, offering a principled path toward simulation models integrate behavior, beliefs, and epidemic outcomes.

\textbf{Our contributions}
    We introduce a graph-based forecasting framework that combines trend similarity graphs with GraphSAGE, and systematically compare different graph construction strategies to demonstrate their critical impact on forecasting accuracy across multiple signals including epidemics, behavior, and beliefs.
    We show that the framework provides interpretability by revealing influential signals and dependencies, offering insights that can inform simulation models for public health.

\begin{figure*}[htbp]
    \centering
    \includegraphics[width=0.8\textwidth]{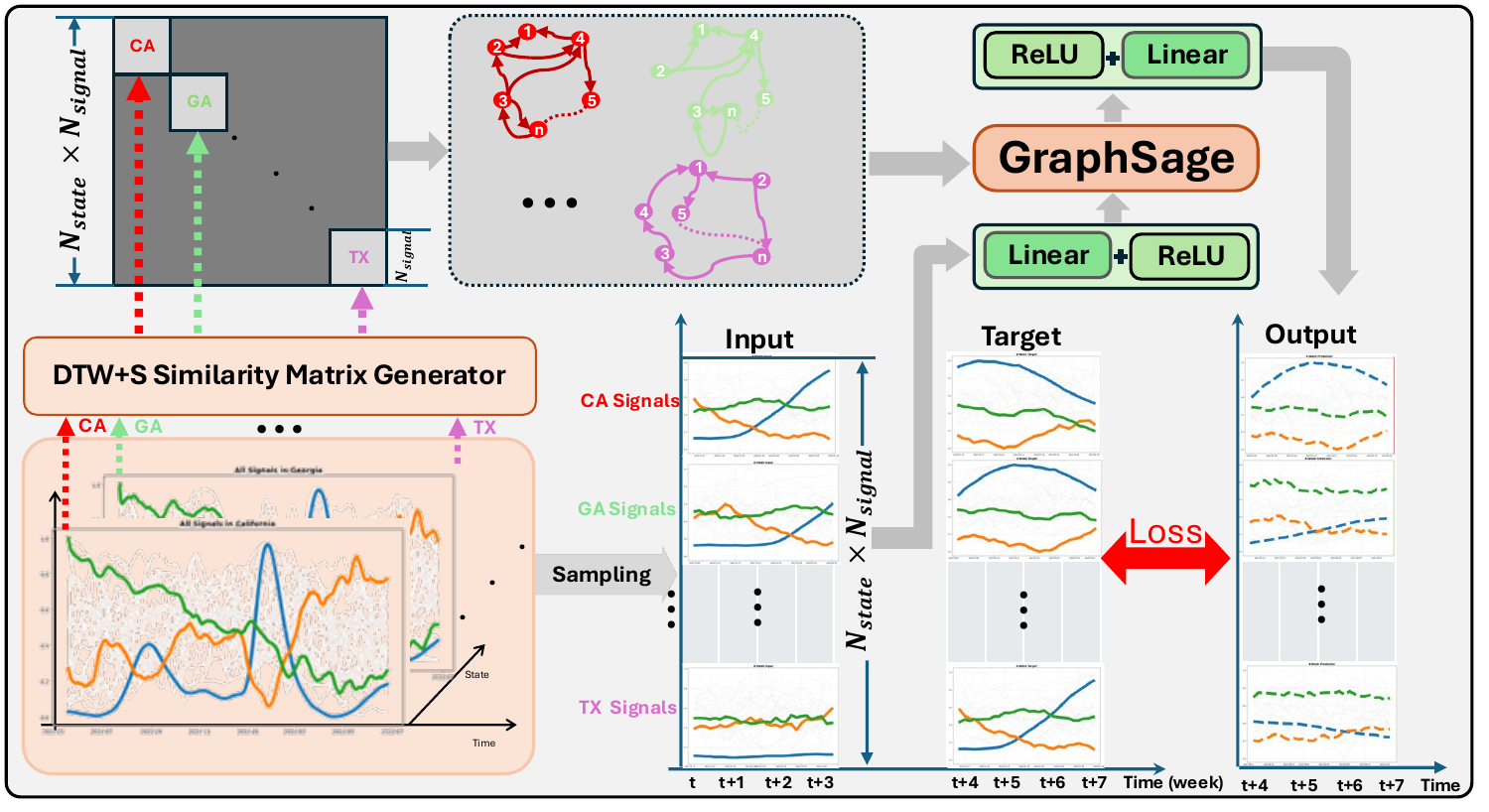}
    \caption{Overview of the TrendGNN pipeline. The dark gray block denotes a block-diagonal matrix, where each light gray sub-block on the diagonal is an $N_{\text{signal}} \times N_{\text{signal}}$ similarity matrix for one state (computed using DTW+S). Off-diagonal regions are set to zero, and stacking across $N_{\text{state}}$ yields a final square matrix of size $(N_{\text{signal}} \times N_{\text{state}})^2$. The Input, Output, and Target in the diagram each have size $(N_{\text{state}} \times N_{\text{signal}}) \times \text{window}$, where the window length is four weeks as shown.}
    \label{fig:method}
\end{figure*}

\section{Background and Related Works}

\textbf{Incorporating Human Behavior} Existing works that combine epidemic and behavior modeling take a simplistic view of both aspects~\cite{kabir2020evolutionary,agusto2022isolate,kwuimy2020nonlinear,kordonis2022dynamic} with one or more drawbacks. (i) They use a simple epidemic model (like SIR) that does not represent the complexities of real-world. (ii) They use simple behavioral models where individuals assess the cost of following interventions at an individual level. (iii) They present simulations without rigorously grounding their assumptions and results with real-world observations. We take a data-driven approach to understand how various signals are related to each other, which can then be used in mechanistic modelling.

\textbf{Epidemic Forecasting}
Early approaches to epidemic forecasting relied on statistical and compartmental models such as ARIMA~\cite{ARIMA} and SEIR~\cite{SEIR}, which capture short-term dynamics but struggle with complex, high-dimensional signals. 
More recent methods apply machine learning and deep learning models, including recurrent neural networks~\cite{RNN} and Transformer~\cite{transformer}, to integrate diverse behavioral and epidemiological data. While these black-box models can achieve competitive accuracy, especially for short-term horizons, they often lack interpretability and do not explicitly model relationships among signals—limitations that motivate graph-based approaches. A number of methods are used in forecast hubs~\cite{bracher2021evaluating,centers2019flusight} but they only target epidemic outcomes (e.g., deaths and hospitalization) and not behavior, which is necessary to understand the effect of interventions.

\textbf{Graph-based Time Series Forecasting}
Graph neural networks (GNNs) have recently been applied to multivariate time series forecasting in domains such as traffic, climate, and energy, where graph structures capture dependencies among signals. Methods such as GCN~\cite{GCN} and GAT\cite{GAT} leverage spatial or relational information to improve prediction accuracy beyond purely sequential models. GraphSAGE\cite{graphsage}, in particular, has been used to efficiently aggregate information from local neighborhoods and scale to larger graphs, making it well-suited for dynamic, high-dimensional forecasting tasks. However, existing work often assumes a fixed or externally provided graph, with limited attention to how the graph itself should be constructed from data -- a key focus of our study.

\textbf{Signal Similarity and Graph Construction}
Graph construction is crucial for forecasting performance. Traditional approaches often rely on lagged correlations, which capture delayed dependencies among signals. More recent methods, such as DTW+S (Dynamic Time Warping with Shapelets)~\cite{DTW+S}, emphasize local trend similarity by aligning shapelet-based representations of time series. These strategies highlight the importance of choosing meaningful similarity measures, though their effectiveness in epidemic forecasting remains underexplored.

\begin{table*}[htbp]
    \centering
    \caption{This table reports the prediction errors of different models on 1–4 week-ahead forecasting tasks, measured by Mean Absolute Error (MAE). \textbf{B}, \textbf{DB}, \textbf{H}, and \textbf{TV} are abbreviations for the four categories introduced in \Cref{Method}. \textbf{AVG} denotes the average error across these four categories. For each column, the worst-performing value is highlighted in red, while the best and second-best values are marked in bold and underlined, respectively.}

\resizebox{\textwidth}{!}{%
\begin{tabular}{l|ccccc|ccccc}
\hline
\multirow{2}{*}{MAE}
  & \multicolumn{5}{c|}{1-Week-Ahead}
  & \multicolumn{5}{c}{2-Week-Ahead} \\
\cline{2-11}
  & B & DB & H & TV & \multicolumn{1}{|c|}{AVG} & B & DB & H & TV & \multicolumn{1}{|c}{AVG} \\
\hline
Baseline & \textbf{0.0729} & \textcolor{red}{0.1148} & \underline{0.0501} & \textcolor{red}{0.1108} & 0.0872 & 0.1082 & \textcolor{red}{0.1376} & \underline{0.0898} & \textcolor{red}{0.1306} & \textcolor{red}{0.1166} \\
ARIMA & \underline{0.0741} & \underline{0.1059} & \underline{0.0504} & 0.1046 & \textbf{0.0837} & 0.1098 & 0.1209 & \textbf{0.0894} & 0.1220 & 0.1105 \\
Sage (Random) & 0.0816 & 0.1064 & 0.0620 & 0.1021 & 0.0880 & 0.1117 & 0.1131 & 0.1006 & 0.1084 & 0.1084 \\
Transformer & 0.0780 & 0.1108 & 0.0629 & 0.1069 & \textcolor{red}{0.0896} & 0.1044 & 0.1195 & 0.0912 & 0.1148 & 0.1075 \\
Sage (Full) & \textcolor{red}{0.0837} & \textbf{0.1055} & \textcolor{red}{0.0646} & \underline{0.1013} & 0.0888 & \textcolor{red}{0.1118} & \textbf{0.1125} & \textcolor{red}{0.1014} & 0.1080 & 0.1084 \\
Sage (Lagged) & 0.0792 & 0.1067 & 0.0609 & \textbf{0.1011} & \underline{0.0870} & \underline{0.1035} & \underline{0.1127} & 0.0975 & \textbf{0.1066} & \underline{0.1051} \\
Sage (DTW+S) & 0.0785 & 0.1064 & 0.0622 & \underline{0.1013} & 0.0871 & \textbf{0.1026} & \textbf{0.1125} & 0.0966 & \underline{0.1067} & \textbf{0.1046} \\
\hdashline 
\multirow{2}{*}{MAE}
  & \multicolumn{5}{c|}{3-Week-Ahead}
  & \multicolumn{5}{c}{4-Week-Ahead} \\
\cline{2-11}
  & B & DB & H & TV & \multicolumn{1}{|c|}{AVG} & B & DB & H & TV & \multicolumn{1}{|c}{AVG} \\
\hline
Baseline & 0.1339 & \textcolor{red}{0.1417} & \textcolor{red}{0.1250} & \textcolor{red}{0.1343} & \textcolor{red}{0.1338} & 0.1606 & \textcolor{red}{0.1434} & \textcolor{red}{0.1553} & \textcolor{red}{0.1367} & \textcolor{red}{0.1490} \\
ARIMA & \textcolor{red}{0.1370} & 0.1279 & 0.1233 & 0.1299 & 0.1295 & \textcolor{red}{0.1624} & 0.1333 & 0.1516 & 0.1365 & 0.1459 \\
Transformer & 0.1263 & 0.1200 & 0.1168 & 0.1148 & 0.1195 & 0.1537 & 0.1249 & 0.1414 & 0.1212 & 0.1353 \\
Sage (Random) & 0.1239 & 0.1128 & 0.1212 & 0.1070 & 0.1162 & 0.1315 & 0.1139 & 0.1347 & 0.1077 & 0.1219 \\
Sage (Full) & \underline{0.1225} & 0.1124 & 0.1203 & 0.1072 & 0.1156 & 0.1277 & \underline{0.1132} & 0.1330 & 0.1072 & 0.1203 \\
Sage (Lagged) & \textbf{0.1126} & \textbf{0.1115} & \underline{0.1162} & \underline{0.1055} & \underline{0.1115} & \underline{0.1217} & 0.1135 & \underline{0.1294} & \underline{0.1069} & \underline{0.1179} \\
Sage (DTW+S) & \textbf{0.1126} & \underline{0.1117} & \textbf{0.1160} & \textbf{0.1050} & \textbf{0.1113} & \textbf{0.1194} & \textbf{0.1129} & \textbf{0.1276} & \textbf{0.1059} & \textbf{0.1165} \\
\hline
\end{tabular}%
}

    \label{tab:mae_results}
\end{table*}

\begin{figure}[htbp]
    \centering
    \includegraphics[width=0.48\textwidth]{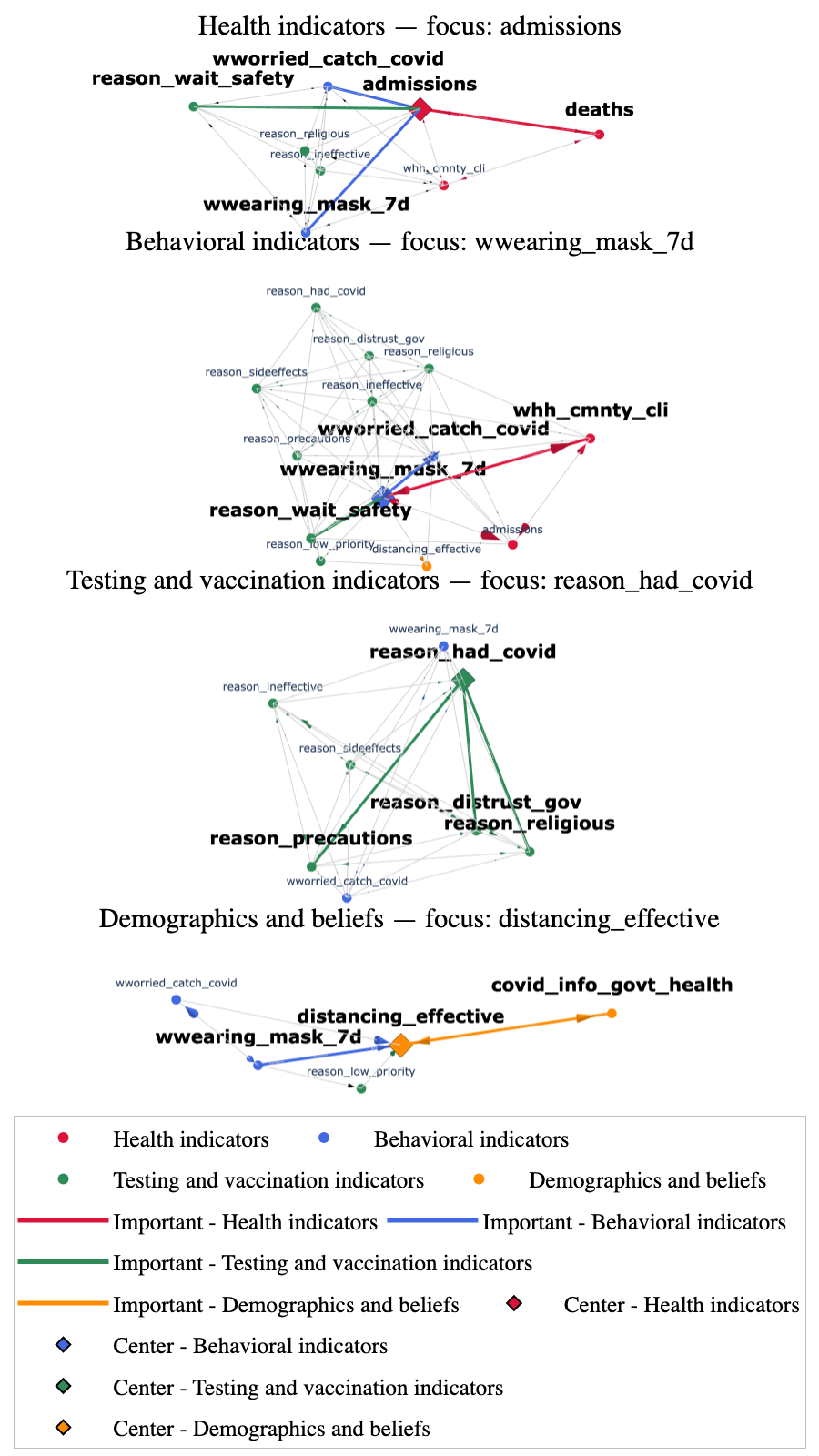}
    \caption{3D visualization of signal graphs across four categories of indicators. Each subplot corresponds to one category. Nodes are colored by category, with the center node (focus signal) highlighted. Thick edges denote important connections.}
    \label{fig:graph_importance}
\end{figure}

\begin{figure}[htbp]
    \centering
    \includegraphics[width=0.43\textwidth]{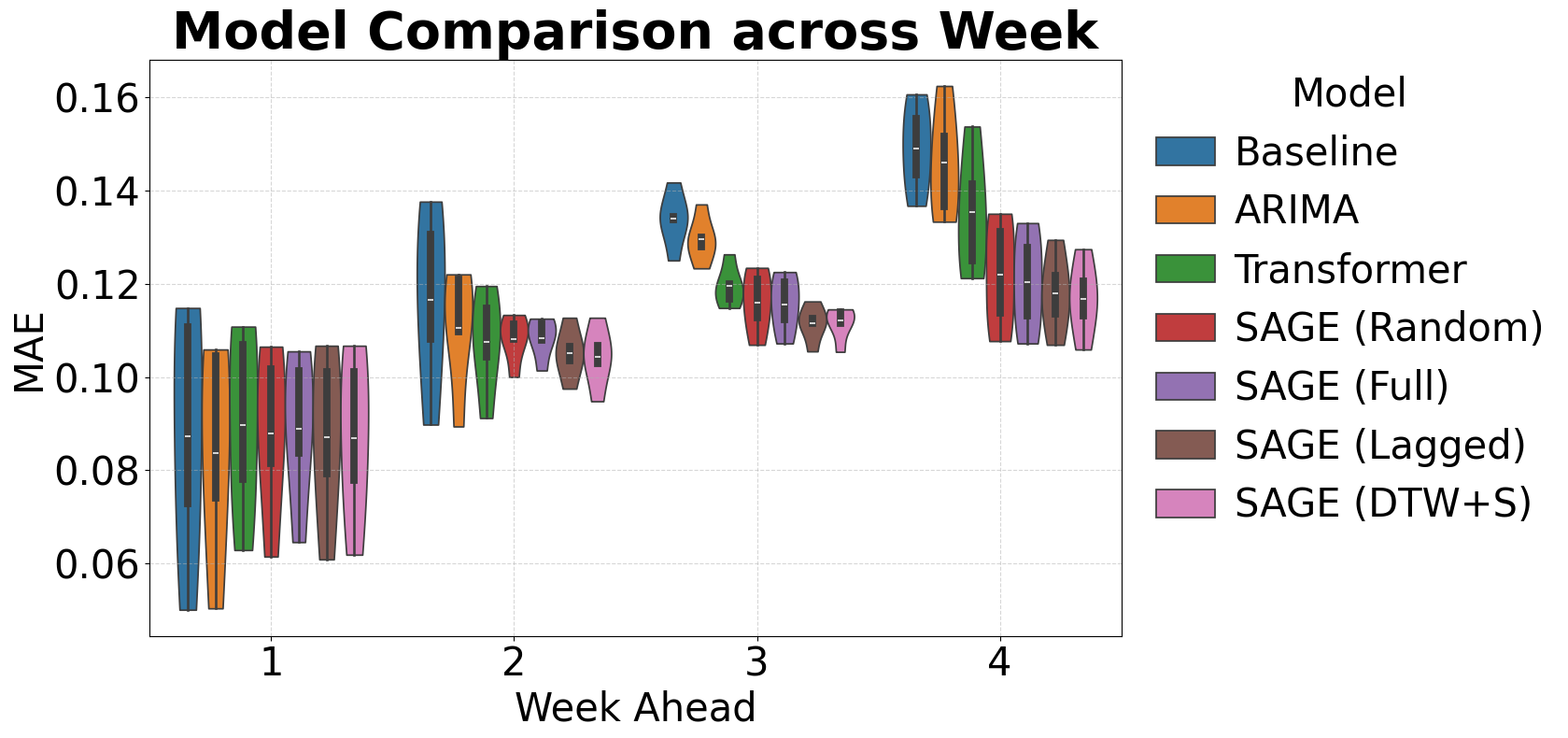}
    \caption{Mean absolute error (MAE) distributions of different models across 1–4 week-ahead forecasting tasks.}
    \label{fig:violin}
\end{figure}

\section{Method}
\label{Method}

Our goal is to build a framework that reveals dependencies across multiple signals and to show its usefulness through a forecasting task. To  demonstrate this we choose a collection of a number of signals available during COVID-19.

\textbf{Dataset}
To validate our approach, we use a subset of the COVID-19 Trends and Impact Survey (CTIS)~\cite{dataset_ctis}. The survey was designed to collect large-scale, timely data on the spread and impact of COVID-19 across the United States.
This subset contains data from \textbf{26 U.S. states} covering \textbf{402 days}, from May 20, 2020 to June 25, 2022. For each state, we include \textbf{26 COVID-19 related signals}, after removing signals that were not fully available in that period. All signals were rescaled using min–max normalization to lie within the interval [0, 1]. %
Each signal belongs to one of four categories (Detailed category grouping and signal explanation are in \cite{dataset_ctis}):

\begin{itemize}[noitemsep, topsep=0pt, leftmargin=*]
    \item \textbf{Health indicators} Derived from self-reported symptoms of influenza-like and COVID-like illness.
    \item \textbf{Behavioral indicators} Reflect individual preventive and mobility behaviors, including mask use, time spent indoors.
    \item \textbf{Testing and vaccination indicators} Represent intentions, barriers, and attitudes to COVID-19 testing and vaccination.
    \item \textbf{Demographics and beliefs} Cover socioeconomic concerns, trust in institutions, and personal beliefs about pandemic.
\end{itemize}

\paragraph{Architecture}
\Cref{fig:method} illustrates the architecture of our TrendGNN workflow. For each state, we generate a similarity matrix using DTW+S and sparsify it by retaining the top five most correlated connections for each signal. The resulting matrices are aggregated into a block-diagonal structure, with each block corresponding to one state. This block-diagonal matrix is then used to construct a unified graph encompassing all nodes across states. Finally, the constructed graph is combined with data sampled using a rolling window of size four and fed into the network for training.

\paragraph{Retrospective Training Setup}

{For the training and evaluation, we adopt a rolling-window strategy to split the dataset into training and testing segments -- this mimics the real-world setup, where every week we are interested in forecasting the outcomes over the next weeks based on the data available so far. Specifically,  $\tau_{\text{start}}$ indicates the number of initial weeks used for training. For example, if $\tau_{\text{start}} = 20$, the model is trained on the first 20 weeks of data and evaluated on the subsequent 4 weeks.
As $\tau_{\text{start}}$ increases, the training window gradually expands while the test set moves forward accordingly.}

\section{Experiments}
Our goal is to show that having a graph of temporal signals improves our ability to forecast all the signals. Specifically, we wish to demonstrate that (1) graph-based methods are better (particularly for longer horizons), and (2) having a ``good'' graph of signals is helpful.  We chose forecasting of a four-week horizon, which is typical in epidemic forecasting~\cite{bracher2021evaluating}. 

\textbf{Baselines} We used several baselines to compare our approach against. Consistent with our approach, all models take the past four weeks of data as input and predict the subsequent four weeks. In addition, to reduce the influence of randomness, each model was trained five times, and the final results are presented as the average value.

\textbf{Non-GNN Methods.} The following baselines do not use the graph obtained form the signals:
(i) \textit{Flat Baseline:} uses the most recent week’s values as predictions for each task;
(ii)\textit{ARIMA:} a traditional statistical model widely used in time-series forecasting; and
(iii) \textit{Transformer:} a modern deep learning approach that captures long-range temporal dependencies.

\textbf{GNN Methods. } To identify the impact of the graph, we construct the graphs with the following strategies:
(i) \textit{Random:} for each state, we generate a directed graph among its 26 signals by randomly assigning edges, ensuring that each node has on average five outgoing edges; 
(ii) \textit{Fully Connected:} for each state, we construct a fully connected directed graph among the 26 signals;
(iii) \textit{Lagged Correlation:} similar to our approach except that DTW+S is replaced by lagged correlation; and
(iv) \textit{DTW+S:} our proposed approach.

\textbf{Results.}
Our experiments demonstrate that graph-based models consistently outperform both traditional models and simple baselines, particularly for medium- to long-term horizons (2–4 weeks) in ~\Cref{tab:mae_results}. Among graph constructions, semantically meaningful approaches such as DTW+S and lagged correlation yield clear gains over random or fully connected graphs. While ARIMA performs slightly better for the 1-week-ahead case, the best and second-best results across 2–4 week forecasts are consistently obtained with DTW+S and lagged-correlation-based graphs. DTW+S is especially effective, as it captures temporal misalignments and non-linear similarities, which are common in COVID-related indicators, while lagged correlation is limited to linear dependencies with fixed shifts. The results highlight that performance gains are driven by leveraging meaningful relational structures.

\section{Graph Explanation}

We envision that our framework will assist with designing models that better capture the underlying mechanisms. As an early step, it is important to identify which relationships are critical in driving the behavior of the signals. While we used DTW+S to select top relationships based on trends,  we use the CF-GNNExplainer~\cite{cfgnnexplainer} framework on our forecasting model to further identify which signals are critical in impacting the predictive behavior of the GNNs. Unlike the original formulation, which identifies the minimal edge changes that flip a node classification, our variant seeks the smallest perturbations that brings the MAE at or below the MAE using random graph. For each signal, we applied CF-GNNExplainer~\cite{cfgnnexplainer} across all states and weeks. We then counted how often the most important edges appeared for that signal and selected the relatively most critical edges as the final results. In~\Cref{fig:graph_importance}, for each category we chose one representative signal and visualized a subgraph from the DTW+S-constructed graph together with its most important edge.

\noindent\textbf{Health indicators}, \textit{hospitalization} were most influenced by mortality outcomes, perceptions of vaccine safety, and worries about catching COVID-19, showing how both epidemiological severity and behavioral attitudes influence admissions.

\noindent\textbf{Behavioral indicators}, \textit{mask-wearing} was strongly shaped by safety concerns, local illness reports, and infection worries, highlighting its dual role as both a preventive action and a proxy for perceived community risk.

\noindent\textbf{Testing and vaccination indicators}, \textit{beliefs about not needing a vaccine due to prior infection} were primarily driven by religious reasons, distrust of government, and alternative precautionary attitudes, revealing how prior experience and beliefs drive hesitancy.

\noindent\textbf{Demographics and beliefs}, perceived \textit{effectiveness of distancing} was largely influenced by protective behaviors and trust in government information, illustrating the interplay between institutional trust and adherence to preventive measures.

\section{Conclusion}

Beyond accuracy, the graph view highlights which signals are more predictable and reveals the dependencies that drive improvements, offering a more interpretable perspective compared to black-box models. These findings underline the importance of constructing meaningful, temporally aligned graphs rather than relying on arbitrary connectivity. Future directions include incorporating lag-aware causal structures to disentangle true drivers from correlations, as well as enhancing edge- and subgraph-level interpretability to support policy decisions and simulation modeling of behavioral and epidemic dynamics.

\vfill\pagebreak
\clearpage
\section*{Acknowledgements}
This work is supported by the Centers for Disease Control and Prevention's (CDC) Center for Forecasting and Outbreak Analytics (CFA) through the Insight Net award.
\label{sec:refs}
\bibliographystyle{IEEEbib}
\bibliography{references}
\clearpage
\appendix

\section{Explanation}
\label{appendix_explanation}
\subsection{Dataset Explanation}
\begin{enumerate}
    \item \textbf{new\_time\_value:} Date in format (Year-Month-Day).
    \item \textbf{smoothed\_waccept\_covid\_vaccine\_no\_appointment:} Estimated percentage of respondents who would definitely or probably choose to get vaccinated, if a vaccine were offered to them today, among respondents who have not yet been vaccinated and do not have an appointment to do so.
    \item \textbf{smoothed\_wbelief\_created\_small\_group:} Estimated percentage of people who believe that the statement ``COVID-19 was deliberately created by a small group of people who secretly manipulate world events'' is definitely or probably true.
    \item \textbf{smoothed\_wbelief\_distancing\_effective:} Estimated percentage of respondents who believe that social distancing is either very or moderately effective for preventing the spread of COVID-19.
    \item \textbf{smoothed\_wbelief\_govt\_exploitation:} Estimated percentage of people who indicate that the statement ``The COVID-19 pandemic is being exploited by the government to control people'' is definitely or probably true.
    \item \textbf{smoothed\_wdelayed\_care\_cost:} Estimated percentage of respondents who have ever delayed or not sought medical care in the past year because of cost.
    \item \textbf{smoothed\_wdontneed\_reason\_dont\_spend\_time:} Estimated percentage of respondents who say they don’t need to get a COVID-19 vaccine because they don’t spend time with high-risk people, among respondents who provided at least one reason for why they believe a COVID-19 vaccine is unnecessary.
    \item \textbf{smoothed\_wdontneed\_reason\_had\_covid:} Estimated percentage of respondents who say they don’t need to get a COVID-19 vaccine because they already had the illness, among respondents who provided at least one reason for why they believe a COVID-19 vaccine is unnecessary.
    \item \textbf{smoothed\_wdontneed\_reason\_precautions:} Estimated percentage of respondents who say they don’t need to get a COVID-19 vaccine because they will use other precautions, such as a mask, instead, among respondents who provided at least one reason for why they believe a COVID-19 vaccine is unnecessary.
    \item \textbf{smoothed\_whesitancy\_reason\_cost:} Estimated percentage of respondents who say they are hesitant to get vaccinated because they are worried about the cost, among respondents who answered ``Yes, probably'', ``No, probably not'', or ``No, definitely not'' when asked if they would get vaccinated if offered (item V3). This series of items was shown to respondents starting in Wave 8.
    \item \textbf{smoothed\_whesitancy\_reason\_distrust\_gov:} Estimated percentage of respondents who say they are hesitant to get vaccinated because they don’t trust the government, among respondents who answered ``Yes, probably'', ``No, probably not'', or ``No, definitely not'' when asked if they would get vaccinated if offered (item V3). This series of items was shown to respondents starting in Wave 8.
    \item \textbf{smoothed\_whesitancy\_reason\_ineffective:} Estimated percentage of respondents who say they are hesitant to get vaccinated because they don’t know if a COVID-19 vaccine will work, among respondents who answered ``Yes, probably'', ``No, probably not'', or ``No, definitely not'' when asked if they would get vaccinated if offered (item V3). This series of items was shown to respondents starting in Wave 8.
    \item \textbf{smoothed\_whesitancy\_reason\_low\_priority:} Estimated percentage of respondents who say they are hesitant to get vaccinated because they think other people need it more than they do, among respondents who answered ``Yes, probably'', ``No, probably not'', or ``No, definitely not'' when asked if they would get vaccinated if offered (item V3). This series of items was shown to respondents starting in Wave 8.
    \item \textbf{smoothed\_whesitancy\_reason\_religious:} Estimated percentage of respondents who say they are hesitant to get vaccinated because it is against their religious beliefs, among respondents who answered ``Yes, probably'', ``No, probably not'', or ``No, definitely not'' when asked if they would get vaccinated if offered (item V3). This series of items was shown to respondents starting in Wave 8.
    \item \textbf{smoothed\_whesitancy\_reason\_sideeffects:} Estimated percentage of respondents who say they are hesitant to get vaccinated because they are worried about side effects, among respondents who answered ``Yes, probably'', ``No, probably not'', or ``No, definitely not'' when asked if they would get vaccinated if offered (item V3). This series of items was shown to respondents starting in Wave 8.
    \item \textbf{smoothed\_whesitancy\_reason\_wait\_safety:} Estimated percentage of respondents who say they are hesitant to get vaccinated because they want to wait to see if the COVID-19 vaccines are safe, among respondents who answered ``Yes, probably'', ``No, probably not'', or ``No, definitely not'' when asked if they would get vaccinated if offered (item V3). This series of items was shown to respondents starting in Wave 8.
    \item \textbf{smoothed\_whh\_cmnty\_cli:} Estimated percentage of people reporting illness in their local community, as described in \href{https://cmu-delphi.github.io/delphi-epidata/api/covidcast-signals/fb-survey.html#estimating-community-cli}{this link}, including their household.
    \item \textbf{smoothed\_wrace\_treated\_fairly\_healthcare:} Estimated percentage of respondents who somewhat or strongly agree that people of their race are treated fairly in a healthcare setting.
    \item \textbf{smoothed\_wspent\_time\_indoors\_1d:} Estimated percentage of respondents who ``spent time indoors with someone who isn’t currently staying with you'' in the past 24 hours.
    \item \textbf{smoothed\_wtrust\_covid\_info\_friends:} Estimated percentage of respondents who trust friends and family to provide accurate news and information about COVID-19.
    \item \textbf{smoothed\_wtrust\_covid\_info\_govt\_health:} Estimated percentage of respondents who trust government health officials to provide accurate news and information about COVID-19.
    \item \textbf{smoothed\_wtrust\_covid\_info\_religious:} Estimated percentage of respondents who trust religious leaders to provide accurate news and information about COVID-19.
    \item \textbf{smoothed\_wwearing\_mask\_7d:} Estimated percentage of people who wore a mask for most or all of the time while in public in the past 7 days; those not in public in the past 7 days are not counted.
    \item \textbf{smoothed\_wworried\_catch\_covid:} Estimated percentage of respondents worrying either a great deal or a moderate amount about catching COVID-19.
    \item \textbf{deaths\_incidence\_num:} Number of new confirmed deaths due to COVID-19, daily.
    \item \textbf{confirmed\_admissions\_covid\_1d\_7dav:} Sum of adult and pediatric confirmed COVID-19 hospital admissions occurring each day.
    \item \textbf{smoothed\_wpublic\_transit\_1d:} Estimated percentage of respondents who ``used public transit'' in the past 24 hours.
\end{enumerate}

\subsection{Signal Grouping}

The 26 signal in our dataset can be grouped into four major categories:

\textbf{Health indicators.}  
whh\_cmnty\_cli, deaths, admissions.

\textbf{Behavioral indicators.}  
wwearing\_mask\_7d, 

wspent\_time\_indoors\_1d, wpublic\_transit,

wworried\_catch\_covid.

\textbf{Testing and vaccination indicators.}  
vaccine\_no\_appointment, reason\_dont\_spend\_time, reason\_had\_covid, reason\_precautions, reason\_cost, reason\_distrust\_gov, reason\_ineffective, reason\_low\_priority, reason\_religious, reason\_sideeffects, reason\_wait\_safety.

\textbf{Demographics and beliefs.}  
created\_small\_group, distancing\_effective, govt\_exploitation, care\_cost, treated\_fairly\_healthcare, covid\_info\_friends, covid\_info\_govt\_health, covid\_info\_religious.

\begin{figure*}[htbp]
    \centering
    \includegraphics[width=1.0\textwidth]{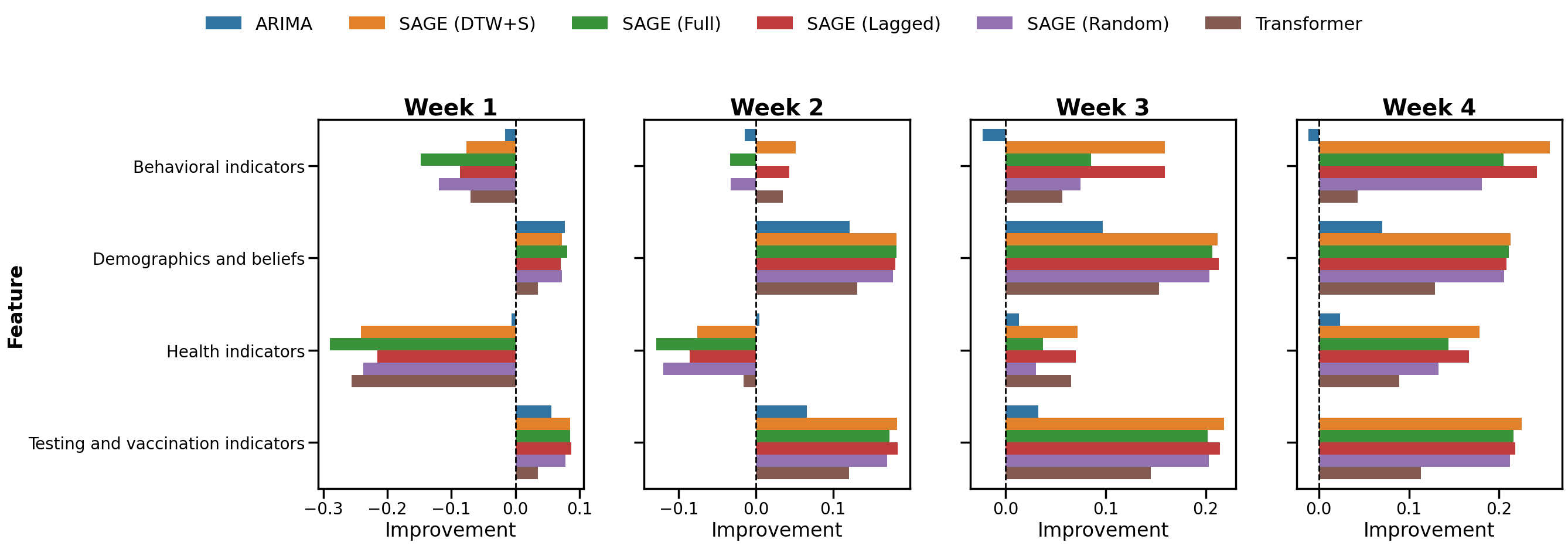}
    \caption{Distribution of relative improvements in MAE for different models over the baseline across 1–4 week-ahead forecasting tasks. The x-axis shows relative improvement, the y-axis shows signal categories, and colors denote models. Positive values indicate gains over the baseline, negative values indicate degradation. }
    \label{fig:improvement}
\end{figure*}

\begin{table*}[htbp]
    \centering
    \caption{Category-level forecasting performance. Mean Absolute Error (MAE) $\pm$ standard deviation, aggregated over signals within each category. The worst-performing models are highlighted in \textcolor{red}{red}, while the best and second-best results are indicated by \textbf{bold} and \underline{underline}, respectively.}
    \resizebox{\textwidth}{!}{%
\begin{tabular}{l|ccccc|ccccc}
\hline
\multirow{2}{*}{MAE}
  & \multicolumn{5}{c|}{1-Week-Ahead}
  & \multicolumn{5}{c}{2-Week-Ahead} \\

\cline{2-11}
  & B & DB & H & TV&\multicolumn{1}{|c|}{AVG}& B & DB & H & TV &\multicolumn{1}{|c}{AVG} \\
\hline
Baseline & \textbf{0.0729 ± NA} & \textcolor{red}{0.1148 ± NA} & \underline{0.0501 ± NA} & \textcolor{red}{0.1108 ± NA} & 0.0872 ± NA & 0.1082 ± NA & \textcolor{red}{0.1376 ± NA} & \underline{0.0898 ± NA} & \textcolor{red}{0.1306 ± NA} & \textcolor{red}{0.1166 ± NA} \\
ARIMA & \underline{0.0741 ±NA} & \underline{0.1059 ±NA} & \underline{0.0504 ±NA} & 0.1046 ± NA & \textbf{0.0837 ±NA} & 0.1098 ±NA & 0.1209 ±NA & \textbf{0.0894 ±NA} & 0.1220 NA & 0.1105 ± NA \\
Sage (Random) & 0.0816 ± 0.0013 & 0.1064 ± 0.0004 & 0.0620 ± 0.0015 & 0.1021 ± 0.0005 & 0.0880 ± 0.0006 & 0.1117 ± 0.0030 & 0.1131 ± 0.0006 & 0.1006 ± 0.0010 & 0.1084 ± 0.0004 & 0.1084 ± 0.0010 \\
Transformer & 0.0780 ± 0.0009 & 0.1108 ± 0.0005 & 0.0629 ± 0.0001 & 0.1069 ± 0.0005 & \textcolor{red}{0.0896 ±0.0004} & 0.1044 ± 0.0013 & 0.1195 ± 0.0009 & 0.0912 ± 0.0006 & 0.1148 ± 0.0005 & 0.1075 ± 0.0005 \\
Sage (Full) & \textcolor{red}{0.0837 ±0.0013} & \textbf{0.1055 ± 0.0005} & \textcolor{red}{0.0646 ±0.0021} & \underline{0.1013 ± 0.0005} & 0.0888 ± 0.0009 & \textcolor{red}{0.1118 ±0.0016} & \textbf{0.1125 ± 0.0006} & \textcolor{red}{0.1014 ± 0.0013} & 0.1080 ± 0.0004 & 0.1084 ± 0.0001 \\
Sage (Lagged) & 0.0792 ± 0.0005 & 0.1067 ± 0.0006 & 0.0609 ± 0.0022 & \textbf{0.1011 ± 0.0006} & \underline{0.0870 ± 0.0008} & \underline{0.1035 ± 0.0019} & \underline{0.1127 ± 0.0003} & 0.0975 ± 0.0031 & \textbf{0.1066 ± 0.0002} & \underline{0.1051 ± 0.0008} \\
Sage (DTW+S) & 0.0785 ± 0.0019 & 0.1064 ± 0.0003 & 0.0622 ± 0.0017 & \underline{0.1013 ± 0.0007} & 0.0871 ± 0.0007 & \textbf{0.1026 ± 0.0025} & \textbf{0.1125 ± 0.0011} & 0.0966 ± 0.0029 & \underline{0.1067 ± 0.0008} & \textbf{0.1046 ± 0.0004} \\
\hdashline 
\multirow{2}{*}{MAE}
  & \multicolumn{5}{c|}{3-Week-Ahead}
  & \multicolumn{5}{c}{4-Week-Ahead} \\

\cline{2-11}
  & B & DB & H & TV&\multicolumn{1}{|c|}{AVG}& B & DB & H & TV &\multicolumn{1}{|c}{AVG} \\
\hline
Baseline & 0.1339 ± NA & \textcolor{red}{0.1417 ± NA} & \textcolor{red}{0.1250 ±NA} & \textcolor{red}{0.1343 ±NA} & \textcolor{red}{0.1338 ± NA} & 0.1606 ± NA & \textcolor{red}{0.1434 ± NA} & \textcolor{red}{0.1553 ± NA} & \textcolor{red}{0.1367 ± NA} & \textcolor{red}{0.1490 ± NA} \\
ARIMA & \textcolor{red}{0.1370 ± NA} & 0.1279 ± NA & 0.1233 ± NA & 0.1299 ± NA & 0.1295 ± NA & \textcolor{red}{0.1624 ± NA} & 0.1333 ± NA & 0.1516 ± NA & 0.1365 ± NA & 0.1459 ± NA \\
Transformer & 0.1263 ± 0.0012 & 0.1200 ± 0.0008 & 0.1168 ± 0.0035 & 0.1148 ± 0.0002 & 0.1195 ± 0.0013 & 0.1537 ± 0.0025 & 0.1249 ± 0.0007 & 0.1414 ± 0.0024 & 0.1212 ± 0.0009 & 0.1353 ± 0.0012 \\
Sage (Random) & 0.1239 ± 0.0050 & 0.1128 ± 0.0007 & 0.1212 ± 0.0021 & 0.1070 ± 0.0005 & 0.1162 ± 0.0015 & 0.1315 ± 0.0039 & 0.1139 ± 0.0007 & 0.1347 ± 0.0013 & 0.1077 ± 0.0012 & 0.1219 ± 0.0008 \\
Sage (Full) & \underline{0.1225 ± 0.0039} & 0.1124 ± 0.0009 & 0.1203 ± 0.0017 & 0.1072 ± 0.0005 & 0.1156 ± 0.0011 & 0.1277 ± 0.0026 & \underline{0.1132 ± 0.0007} & 0.1330 ± 0.0034 & 0.1072 ± 0.0005 & 0.1203 ± 0.0015 \\
Sage (Lagged) & \textbf{0.1126 ± 0.0030} & \textbf{0.1115 ± 0.0007} & \underline{0.1162 ± 0.0030} & \underline{0.1055 ± 0.0006} & \underline{0.1115 ± 0.0012} & \underline{0.1217 ± 0.0023} & 0.1135 ± 0.0010 & \underline{0.1294 ± 0.0021} & \underline{0.1069 ± 0.0007} & \underline{0.1179 ± 0.0004} \\
Sage (DTW+S) & \textbf{0.1126 ± 0.0033} & \underline{0.1117 ± 0.0008} &  \textbf{0.1160 ± 0.0022} &  \textbf{0.1050 ± 0.0007} &  \textbf{0.1113 ± 0.0007} & \textbf{0.1194 ± 0.0044} &  \textbf{0.1129 ± 0.0006} &  \textbf{0.1276 ± 0.0048} &  \textbf{0.1059 ± 0.0008} &  \textbf{0.1165 ± 0.0017} \\
\hline
\end{tabular}%
}

    \label{tab:mae_results_std}
\end{table*}

\end{document}